\documentclass[sigconf]{acmart}

\usepackage[ruled]{algorithm2e}
\usepackage{multirow}
\usepackage{subfigure}
\usepackage{enumitem}
\usepackage{sistyle}
\SIthousandsep{,}
\usepackage{makecell}
\usepackage{multirow}
\usepackage{enumitem}
\usepackage{filecontents}
\usepackage{bbm}
\usepackage{pgfplots}
\usetikzlibrary{patterns}

\newcommand\blfootnote[1]{%
  \begingroup
  \renewcommand\thefootnote{}\footnotetext{#1}%
  \addtocounter{footnote}{-1}%
  \endgroup
}

\AtBeginDocument{%
  \providecommand\BibTeX{{%
    \normalfont B\kern-0.5em{\scshape i\kern-0.25em b}\kern-0.8em\TeX}}}

\copyrightyear{2022} 
\acmYear{2022} 
\setcopyright{acmcopyright}
\acmConference[SIGIR '22]{Proceedings of the 45th International ACM SIGIR Conference on Research and Development in Information Retrieval}{July 11--15, 2022}{Madrid, Spain}
\acmBooktitle{Proceedings of the 45th International ACM SIGIR Conference on Research and Development in Information Retrieval (SIGIR '22), July 11--15, 2022, Madrid, Spain}
\acmPrice{15.00}
\acmDOI{10.1145/3477495.3531827}
\acmISBN{978-1-4503-8732-3/22/07}

\makeatletter

\begin{document}
\fancyhead{}

\title{GERE: Generative Evidence Retrieval for Fact Verification}

\author{Jiangui Chen, Ruqing Zhang, Jiafeng Guo$^{*}$, Yixing Fan, \lowercase{and} Xueqi Cheng}
\affiliation{%
  \institution{
    CAS Key Lab of Network Data Science and Technology, Institute of Computing Technology, \\ Chinese Academy of Sciences, Beijing, China\\
    University of Chinese Academy of Sciences, Beijing, China\\}
  \city{}
  \country{}
}
\email{{chenjiangui18z, zhangruqing, guojiafeng, fanyixing, cxq}@ict.ac.cn}

\begin{abstract}

\blfootnote{$^{*}$ Jiafeng Guo is corresponding author}

Fact verification (FV) is a challenging task which aims to verify a claim using multiple evidential sentences from trustworthy corpora, e.g., Wikipedia.   
Most existing approaches follow a three-step pipeline framework, including document retrieval, sentence retrieval and claim verification. 
High-quality evidences provided by the first two steps are the foundation of the effective reasoning in the last step. 
Despite being important, high-quality evidences are rarely studied by existing works for FV, which often adopt the off-the-shelf models to retrieve relevant documents and sentences in an ``index-retrieve-then-rank'' fashion. 
This classical approach has clear drawbacks as follows: i) a large document index as well as a complicated search process is required, leading to considerable memory and computational overhead; ii) independent scoring paradigms fail to capture the interactions among documents and sentences in ranking; 
iii) a fixed number of sentences are selected to form the final evidence set. In this work, we propose \textit{GERE}, the first system that retrieves  evidences in a generative fashion, i.e., generating the document titles as well as evidence sentence identifiers.  
This enables us to mitigate the aforementioned technical issues since: 
i) the memory and computational cost is greatly reduced because the document index is eliminated and the heavy ranking process is replaced by a light generative process; 
ii) the dependency between documents and that between sentences could be captured via sequential generation process; 
iii) the generative formulation allows us to dynamically select a precise set of relevant evidences for each claim. 
The experimental results on the FEVER dataset show that GERE achieves significant improvements over the state-of-the-art baselines, with both time-efficiency and memory-efficiency. 

\end{abstract}

\ccsdesc[500]{Information systems~Retrieval models and ranking}
\ccsdesc[500]{Computing methodologies~Natural language processing}

\keywords{Fact Verification; Evidence Retrieval; Generative Retrieval}

\maketitle

\section{Introduction}

With the growing online contents with false information, such as fake news, political deception and online rumors, how to automatically ``fact check'' the integrity of information is urgently needed for our society. 
Hence, many recent research efforts have been devoted to the fact verification (FV) task, which targets to automatically verify the truthfulness of a textual claim using multiple evidential sentences from trustworthy corpora, e.g., Wikipedia. 

The majority of existing approaches adopts a three-step pipeline framework \cite{thorne2018fever, gear, kgat, nsmn, chakrabarty2018robust}, where document retrieval and sentence retrieval component are employed to retrieve relevant documents and sentences respectively to provide evidences to the claim verification component. 
In essence, high-quality evidences are the foundation of claim verification to support effective determination of the veracity of claims. 
Existing methods for FV mainly make the claim verification the primary concern and directly adopt the off-the-shelf ``index-retrieve-then-rank'' approaches  for evidence retrieval. 
Unfortunately, such approach has several shortcomings. Firstly, document retrieval needs a large document index to search over the given corpus, which requires considerable memory resources to store the whole data. 
The complicated search process for both the document and sentence retrieval also causes significant computation overhead.   
Secondly, the relevance is modeled independently on each document and sentence, missing cross-candidate interactions and local context information. Finally, a fixed number of top-ranked documents and sentences are selected for the final evidence set, limiting the flexibility in verifying different claims.

Therefore, in this paper, we propose to bypass the explicit retrieval process and introduce \textit{GERE} (for \textit{Generative Evidence REtrieval}), the first system that retrieves evidences in a generative way. 
Specifically, GERE exploits a transformer-based encoder–decoder architecture, pre-trained with a language modeling objective and fine-tuned to generate document titles and evidence sentence identifiers jointly.  
To our knowledge, this is the first attempt to use this generative class of retrieval models for FV.  
Finally, based on the evidences obtained by GERE, we train an existing claim verification model to verify the claim. 
GERE enables us to mitigate the aforementioned technical issues.  
Firstly, the memory footprint is greatly reduced, since the parameters of a sequence-to-sequence model scale linearly with the vocabulary size, not document count. 
The heavy ranking process is replaced with a light generative process, and thus we can skip the time-consuming step of searching over a potentially massive space of options. 
Secondly, GERE considers the dependency information, which contributes to improving the consistency and eliminating duplication among the evidences. Finally, the generative formulation allows us to dynamically decide the number of relevant documents and sentences for different claims.

We conduct experiments on the large-scale benchmark Fact Extraction and VERification (FEVER) dataset. 
Experimental results demonstrate that GERE can significantly outperform the  state-of-the-art baseline systems. 
The results also show that GERE leads to a much smaller memory footprint and time cost.

\section{Related Work}

This section reviews previous studies on fact verification and generative retrieval models. 

\subsection{Fact Verification}
The majority of the most successful FV framework is a three-step pipeline system, i.e., document retrieval, sentence retrieval and claim verification~\cite{feverb, kgat, gear, nsmn, nie2019revealing}. 
For document retrieval, the existing methods can be generally divided into three categories, i.e., mention-based methods \cite{hanselowski2018ukp, chernyavskiy2019extract, kgat, soleimani2020bert, zhao2020transformer-xh}, keyword-based methods \cite{nsmn, ma2019sentence, nie2019revealing} and feature-based methods \cite{hidey2018team, ucl_mrg, tokala2019attentivechecker}.
However, these methods generally needs a large document index to search over the corpus, requiring a large memory footprint~\cite{genre}. 
For sentence retrieval, three types of approaches are usually used, including traditional probabilistic ranking models  \cite{thorne2018fever, portelli2020distilling, chernyavskiy2019extract}, neural ranking models  \cite{hanselowski2018ukp, luken2018qed, nie2019revealing, nsmn}, and pre-trained models \cite{soleimani2020bert, kgat, subramanian2020hierarchical, zhao2020transformer-xh}.
However, these approaches model the relevance independently, which lack of flexibility to select specific number of sentences for different claims. 
For claim verification, most recent studies formulate it as a graph reasoning task and pre-trained language models like BERT~\cite{bert} have been widely used \cite{li2019several, gear, soleimani2020bert}.
\citet{yin2018twowingos} proposed a supervised training method named TwoWingOS to jointly conduct sentence retrieval and claim verification. 
Without loss of generality, this step heavily depends on the retrieved evidences, i.e., a precise set of evidences could lead to a better verification result~\cite{cedr, kgat}.

\subsection{Generative Retrieval Models}
Recently, generative models have attracted an increasing amount of attention in the information retrieval (IR) field. 
Different from the commonly-adopted ``index-retrieve-then-rank'' blueprint in previous works \cite{nogueira2019passage, guo2016deep}, generative models focus on predicting relevant documents given a query based on a generation model \cite{ahmad2019context, nogueira2019document, nogueira2020document}. 
For example, ~\citet{metzler2021rethinking} envisioned a model-based IR approach that replaces the long-lived "retrieve-then-rank" paradigm into a single consolidated model. 
~\citet{genre} proposed to retrieve entities by generating their unique names in an autoregressive fashion.  
In this work, we make the first attempt to adapt a pre-trained sequence-to-sequence model, which has been shown to retain factual knowledge, to the evidence retrieval task in FV.

\section{Our Approach}

In this section, we present the Generative Evidence REtrieval (GERE), a novel generative framework for evidence retrieval in FV.

\subsection{Model Overview}
Suppose $\mathcal{K}=\{d_0,d_1,\dots\}$ denotes a large-scale text corpus where $d_i$ denotes an individual document. 
$d_i$ is composed of a sequence of sentences, namely $d_i=\{t_i,s_i^0,s_i^1,\dots,s_i^m\}$ with $t_i$ denoting the document title of $d_i$ and each $s_i^j$ denoting the $j$-th sentence in $d_i$. 
Given a claim $c=\{c_1,c_2,\dots,c_O\}$ and a corpus $\mathcal{K}$, GERE aims to find a set of relevant documents $\hat{D} \subset \mathcal{K}$ and evidential sentences $\hat{E} \subset \hat{D}$ in a generative way, such that $\hat{D} = D$ and  $\hat{E} = E$.  
This goal is different from the goal targeted by existing methods, which aim to retrieve documents $\hat{D} \subset \mathcal{K}$ and evidential sentences $\hat{E} \subset \hat{D}$ such that $D \subseteq \hat{D}$ and $E \subseteq \hat{E}$.

\begin{figure}[t]
 \centering
 \includegraphics[scale=0.35]{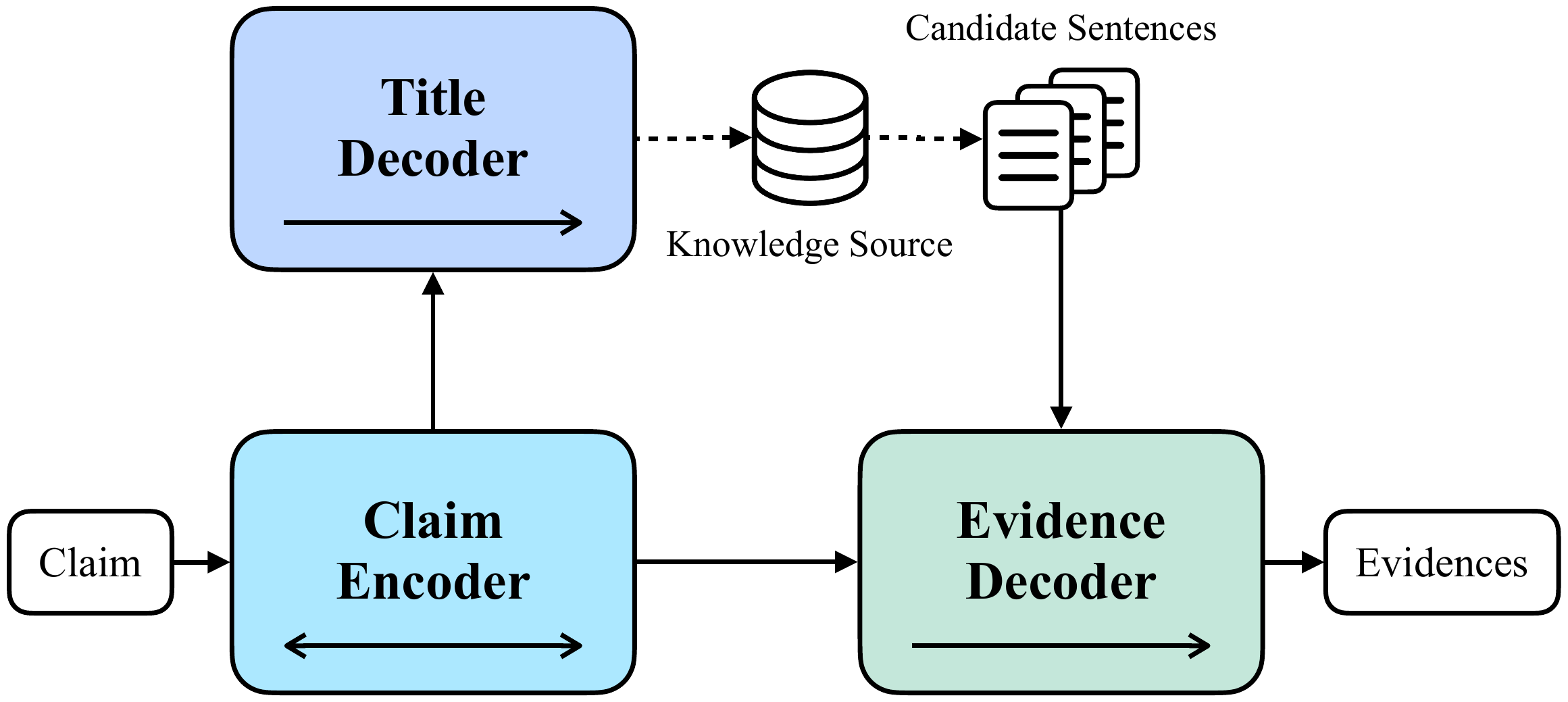}
 \caption{The overview of the GERE framework.}
 \label{fig:framework}
\end{figure}

Basically, the GERE contains the following three dependent components: 
(1) Claim Encoder, a bidirectional encoder to obtain the claim representation; 
(2) Title Decoder, a sequential generation process to produce document titles;  
(3) Evidence Decoder, a sequential generation process to produce evidence sentence identifiers based on the relevant documents. 
The overall architecture of GERE is illustrated in Figure \ref{fig:framework}. 
GERE can be easily adapted to different claim verification models to provide evidences for FV.

\subsection{Claim Encoder}
The goal of the claim encoder is to map the input claim into a
compact vector that can capture its essential topics. 
Specifically, the encoder represents the claim $c$ as a series of hidden vectors, i.e., 
\begin{equation}
    H_{enc} = \text{Encoder} (c_1, c_2, \dots, c_O),
\end{equation}
where $H_{enc}$ denotes the claim representation. 
In this work, we adopt the bidirectional Transformer-based encoder of BART \cite{bart} as the claim encoder due to its superiority in many natural language generation tasks. 

\subsection{Title Decoder}

The goal of the title decoder is to generate a sequence of document titles $[t_1, t_2, \dots, t_{\vert \hat{D} \vert}]$ with respect to $\vert \hat{D} \vert$ relevant documents for each claim. 
Inspired by that ``the relevance of any additional relevant document clearly depends on the relevant documents seen before'' \cite{fuhr2008probability}, the generation of a new title is decided by both the claim and previous generated titles.

Specifically, the title decoder predicts the $n$-th token $w_{m,n}$ in the $m$-th title $t_m$ as follows, 
\begin{equation*}
    p(w_{m,n}|w_{\leq m, <n}, c) = \text{Decoder}(w_{\leq m, <n}, H_{enc}),
\end{equation*}
where we adopt the BART~\cite{bart} decoder as the title decoder. In this way, it is feasible to achieve dynamic predictions of relevant documents for different claims, i.e., a precise set of relevant documents. The title generation objective over the training corpus $\mathcal{D}$ can be further formalized as, 
\begin{equation*}
	\mathcal{L}_{title}=\arg \max_{\theta}\sum_{c\in\mathcal{D}}\sum_{m} \sum_{n} \log p(w_{m,n}|w_{\leq m, <n}, c;\theta),
\end{equation*}
where $\theta$ denotes the model parameters.

At the inference time, we adopt Beam Search \cite{sutskever2014sequence} for token prediction, since it might be prohibitively expensive to compute a score for every document in Wikipedia with $\sim$5M documents. 
Naturally, the generated output might not always be a valid document title if allowing to generate any token from the vocabulary at every decoding step. 
To solve this problem, we employ a constrained Beam Search strategy \cite{genre} to force each generated title to be in a predefined candidate set, i.e., the titles of all the documents in $\mathcal{K}$. 
Specifically, we define our constrain in terms of a prefix tree where nodes are annotated with tokens from the predefined candidate set. 
For each node in the prefix tree, its children indicate all the allowed continuations from the prefix defined traversing the tree from the root to it.
Note when generating a document title (e.g., Wikipedia title), the prefix tree is relatively small and it can be computed and stored into memory in advance.

\subsection{Evidence Decoder}

The evidence decoder is responsible for producing a sequence of evidence sentence identifiers $[e_1, e_2, \dots, e_{\vert \hat{E} \vert}]$ with respect to $\vert \hat{E} \vert$ relevant evidences given the claim and relevant documents. 
In this work, we define the evidence sentence identifier $e_g$ as the unique number denoting the semantic representation of each relevant sentence. 
Specifically, we first identify the relevant documents from $\mathcal{K}$  based on the titles generated by the title decoder. 
Then, we collect all the sentences in relevant documents as the candidate sentence set. 
Finally, we use another BART encoder to encode these sentences to obtain their semantic representations.

Therefore, the evidence decoder predicts the $g$-th sentence identifier  $e_g$ as follows, 
\begin{equation*}
    p(e_g|e_{< g}, c) = \text{Decoder}(e_{< g}, H_{enc}),
\end{equation*}
where we also adopt the BART decoder as the evidence decoder.
Note the embedding vocabulary for the evidence decoder is dynamic and it is composed of the semantic representations of the candidate sentences with respect to different claims.  
Similar to the title decoder, we can not only model the dependency information among evidences, but also achieve a dynamic set of evidences.  
The evidence generation objective over the training corpus $\mathcal{D}$ can be further formalized as, 
\begin{equation*}
	\mathcal{L}_{evidence}=\arg \max_{\beta} \sum_{c\in\mathcal{D}} \sum_{g} \log p(e_g|e_{< g}, c; \beta),
\end{equation*}
where $\beta$ denotes the model parameters. At the inference time, we adopt the greedy decoding strategy due to the small search space.

\subsection{Model Training}

We take advantage of fine-tuning the pre-trained encoder-decoder architecture (i.e., we use BART weights \cite{bart}) and train GERE by maximizing the title generation objective jointly with evidence generation objective, as follow, 
$$\mathcal{L}_{total} =\mathcal{L}_{title} + \mathcal{L}_{evidence}.$$

In this way, the document retrieval and sentence retrieval component can be integrated into a unified framework and trained simultaneously. 
Given a new claim, we can easily employ the learned GERE framework to provide relevant documents as well as evidences to existing claim verification models.

\section{Experiments}

In this section, we conduct experiments to verify the effectiveness of our proposed method.

\subsection{Experimental Settings}

We conduct experiments on the benchmark dataset FEVER ~\cite{thorne2018fever}, which consists of $185,455$ annotated claims with $5,416,537$ Wikipedia documents.  
All claims are classified as SUPPORTS, REFUTES or NOT ENOUGH INFO by annotators. 
For each claim, the average number of relevant documents and evidences is 1.28 and 1.86, respectively.  
The dataset partition is kept the same with \cite{feverb}.

We implement our model in PyTorch based on fairseq library~\footnote[1]{The data and code can be found at \url{https://github.com/Chriskuei/GERE}}. 
We use the BART weights in the large version and optimize the model using Adam \cite{kingma2014adam} with the warmup technique, where the learning rate increases over the first 10\% of batches, and then decays linearly to zero. 
In the training process, we make use of the given order of documents and evidences in FEVER as the ground-truth sequential document titles and evidence identifiers. 
The learning rate is $3e^{-5}$,  the label smoothing rate is 0.1, and the batch size is dynamic with a requirement that the max number of tokens is 4096 for each update. 
The size of the beams is 5. 
All hyper-parameters are tuned using the development set. 

\subsection{Evaluation Metrics}
Following previous works~\cite{nsmn, thorne2018fever, dqn}, for the document retrieval and sentence retrieval, Precision (P), Recall (R) and F1 are used in our experiments. 
Note F1 is our main metric because it can directly show the performance of methods on retrieval of precise documents and evidences. 
For the claim verification, we adopt the official evaluation metrics, i.e., FEVER and Label Accuracy (LA). 
FEVER measures the accuracy with a requirement that the predicted evidences fully covers the ground-true evidences. 
LA measures the accuracy without considering the validity of the retrieved evidences.

\begin{table}[t]
	\caption{Comparisons of the document retrieval performance achieved by  GERE and baselines; $^\ddag$ indicates statistically significant improvements over all the baselines (p-value < 0.05).}
	\label{tab:document}
 \renewcommand{\arraystretch}{1.0}
 \setlength\tabcolsep{10pt}
	\begin{tabular}{lccc}
		\toprule
		& \multicolumn{3}{c}{\textbf{Dev}} \\
		\cmidrule(lr){2-4}
		\textbf{Model} & \textbf{P} & \textbf{R} & \textbf{F1}\\
		\midrule
		BM25 & 14.42 & 66.22 & 23.68 \\
		TF-IDF \cite{nsmn} & 42.83 & 87.45 & 57.50 \\
		\midrule
		UKP-Athene \cite{hanselowski2018ukp} & 35.33 & \textbf{92.51} & 51.13 \\
		\midrule
		KM \cite{nsmn} & 44.90 & 83.30 & 58.35 \\
		NSMN \cite{nsmn} & 52.73 & 88.63 & 66.12 \\
		\midrule
		DPR & 55.42 & 89.35 & 68.41 \\
		RAG & 62.17 & 91.63 & 74.08 \\
		\midrule
		GERE & \textbf{84.43}$^\ddag$ & 78.01 & \textbf{81.10}$^\ddag$ \\
		\bottomrule
	\end{tabular}
\end{table}

\subsection{Results on Document Retrieval}

Following the previous works \cite{nsmn, hanselowski2018ukp}, we choose three types of document retrieval baselines that are widely adopted in FV, including feature-based (i.e., BM25~\cite{robertson2009probabilistic}, TF-IDF~\cite{nsmn}), mention-based (i.e., UKP-Athene~\cite{hanselowski2018ukp}), keyword-based (i.e., Keyword Matching~\cite{nsmn}, NSMN~\cite{nsmn}) methods.  
We also select two dense retrieval methods (i.e., DPR \cite{dpr} and RAG \cite{rag}) for comparison. 
Specifically, documents with top-5 relevance scores are selected in the baselines following the common setting \cite{hanselowski2018ukp, nsmn}. 
We do not truncate the generated sequence of documents in GERE, where 91.24\% claims are provided with less than 5 documents.

As shown in Table ~\ref{tab:document}, we can find that:
(1) GERE significantly outperforms the state-of-the-art methods in terms of F1, demonstrating the superiority of GERE in retrieval of relevant documents. 
(2) The best performance of precision GERE brings comes at the cost of reduced recall compared to baselines. 
The reason might be that GERE generates a preciser but more compact set of documents with respect to each claim. 
That is, GERE leads to a lower number of retrieved documents than baselines, with the average number as 1.91. 
Therefore, the baselines have a higher probability to recall the ground-truth  evidences than our GERE. 

\begin{table}
	\caption{Comparisons of the sentence retrieval performance achieved by  GERE and baselines; $^\ddag$ indicates statistically significant improvements over all the baselines (p-value < 0.05).}
	\label{tab:sentence}
 \renewcommand{\arraystretch}{1.0}
 \setlength\tabcolsep{2.5pt}
	\begin{tabular}{lcccccc}
		\toprule
		& \multicolumn{3}{c}{\textbf{Dev}} & \multicolumn{3}{c}{\textbf{Test}}  \\
		\cmidrule(lr){2-4} \cmidrule(lr){5-7} 
		\textbf{Model} & \textbf{P} & \textbf{R} & \textbf{F1} & \textbf{P} & \textbf{R} & \textbf{F1} \\
		\midrule
		TF-IDF \cite{thorne2018fever} & - & - & 17.20 & 11.28 & 47.87 & 18.26 \\
		ColumbiaNLP \cite{chakrabarty2018robust} & - & 78.04 & - & 23.02 & 75.89 & 35.33 \\
		\midrule
		UKP-Athene \cite{hanselowski2018ukp} & - & 87.10 & - & 23.61 & 85.19 & 36.97 \\
		GEAR \cite{gear} & 24.08 & 86.72 & 37.69 & 23.51 & 84.66 & 36.80 \\
		NSMN \cite{nsmn} & 36.49 & 86.79 & 51.38 & 42.27 & 70.91 & 52.96 \\
		\midrule
		KGAT \cite{kgat} & 27.29 & \textbf{94.37} & 42.34 & 25.21 & \textbf{87.47} & 39.14 \\
		DREAM \cite{zhong2020reasoning} & 26.67 & 87.64 & 40.90 & 25.63 & 85.57 & 39.45 \\
		DQN \cite{dqn} & 54.75 & 79.92 & 64.98 & 52.24 & 77.93 & 62.55 \\
		\midrule
		GERE & \textbf{58.43}$^\ddag$ & 79.61 & \textbf{67.40}$^\ddag$ & \textbf{54.30}$^\ddag$ & 77.16 & \textbf{63.74}$^\ddag$ \\
		\bottomrule
	\end{tabular}
\end{table}

\subsection{Results on Sentence Retrieval}

For sentence retrieval, we adopt several representative models for comparison, including traditional probabilistic ranking models (i.e., TF-IDF~\cite{thorne2018fever}, ColumbiaNLP~\cite{chakrabarty2018robust}), neural ranking models (i.e., UKP-Athene~\cite{hanselowski2018ukp}, GEAR~\cite{gear}, NSMN~\cite{nsmn}) and pre-trained models (i.e., KGAT~\cite{kgat}, DREAM~\cite{zhong2020reasoning}, DQN~\cite{dqn}). 
For the online evidence evaluation, only the first 5 sentences of predicted evidences that the candidate system provides are used for scoring. 
To meet the requirements of the online evaluation, top-$5$ ranked sentences are selected in the baselines and the first $5$ generated sentences are kept in GERE as the evidence set. 
Note that 83.57\% claims are provided with less than 5 sentences in GERE.  

As shown in Table ~\ref{tab:sentence}, we can see that:
(1) Similar to document retrieval, GERE gives the best performance in terms of precision and F1, and performs worse than the baselines in terms of recall for sentence retrieval. 
This is due to that the average number of generated sentences is 2.42 in GERE, i.e., GERE provides more compact but preciser evidence set compared with baselines. 
(2) DQN leverages a post-processing strategy that can also find precise evidences. 
Although it achieves slightly better performance on recall than ours, its precision and F1 are much worse, indicating that post processing is not a optimal way to find precise evidences.

\begin{table}[t]
	\caption{Comparisons of different claim verification models using the evidences obtained from the original paper and that achieved by our GERE.}
	\label{tab:verification}
 \renewcommand{\arraystretch}{1.0}
 \setlength\tabcolsep{6pt}
	\begin{tabular}{lcccc}
		\toprule 
		& \multicolumn{2}{c}{\textbf{Dev}} & \multicolumn{2}{c}{\textbf{Test}}  \\
		\cmidrule(lr){2-3} \cmidrule(lr){4-5} 
		\textbf{Model} & \textbf{LA} & \textbf{FEVER} & \textbf{LA} & \textbf{FEVER} \\
		\midrule
		BERT Concat \cite{kgat} & 73.67 & 68.89 & 71.01 & 65.64 \\
		BERT Concat+GERE & 74.41 & 70.25 & 71.83 & 66.40 \\
		\midrule
		BERT Pair \cite{kgat} & 73.30 & 68.90 & 69.75 & 65.18 \\
		BERT Pair+GERE & 74.59 & 69.92 & 70.33 & 66.51 \\
		\midrule
		GEAR \cite{gear} & 74.84 & 70.69 & 71.60 & 67.10 \\
		GEAR+GERE & 75.96 & 71.88 & 72.52 & 68.34 \\
		\midrule
		GAT \cite{kgat} & 76.13 & 71.04 & 72.03 & 67.56 \\
		GAT+GERE & 77.09 & 72.36 & 72.81 & 69.40 \\
		\midrule
		KGAT \cite{kgat} & 78.29 & 76.11 & 74.07 & 70.38 \\
		KGAT+GERE & \textbf{79.44} & \textbf{77.38} & \textbf{75.24} & \textbf{71.17} \\
		\bottomrule
	\end{tabular}
\end{table}

\subsection{Results on Claim Verification}

To verify the effectiveness of the evidences obtained by GERE, we choose several advanced claim verification models for comparison, including BERT-based models (i.e., BERT-concat~\cite{kgat} and BERT-pair~\cite{kgat}), and graph-based models (i.e., GEAR~\cite{gear}, GAT~\cite{kgat} and KGAT~\cite{kgat}). 
Specifically, these models are provided by the evidences obtained from the original paper and that achieved by GERE, respectively. 
Note we select the first 5 sentences generated by GERE as the evidences for these models to ensure a fair comparison. 

As shown in Table~\ref{tab:verification}, we can observe that:
(1) All the claim verification models using the evidences obtained by our GERE significantly outperform the corresponding original versions. 
This result demonstrates the superiority of our method in retrieval of high-quality evidences. 
Modeling the dependency among documents and dependency among sentences does help improve the quality of evidences and contribute to claim verification. 
(2) By conducting further analysis, we find that the provided evidences in   original papers generally contain conflicting pieces, some of which support the claim while the other refute. 
In this way, the verification process will be misled. 
For GERE, it can provide preciser evidences and the consistency between evidences are improved for different claims. 
The analysis again indicates that the generative formulation is helpful in improving the quality of evidences.

\begin{table}[t]
	\caption{Comparisons on the memory footprint, the number of model parameters and inference time.}
	\label{tab:memory}
 \renewcommand{\arraystretch}{1.0}
 \setlength\tabcolsep{8pt}
	\begin{tabular}{lccc}
		\toprule 
		\textbf{Model} & \textbf{Memory} & \textbf{Parameter} & \textbf{Time} \\ 
		\midrule
		NSMN & 19GB & 502M & 28.51ms\\ 
		DPR & 70.9GB & 220M & 13.89ms \\
		RAG & 40.4GB & 626M & 9.46ms \\
		\midrule
		\textbf{GERE} & \textbf{2.1GB} & \textbf{581M} & \textbf{5.35ms} \\ 
		\bottomrule 
	\end{tabular}
\end{table}

\subsection{Memory and Inference Efficiency}

In general, the document retrieval step constitutes the major part of the memory and computation cost for FV.  
Therefore, we evaluate the inference time of document retrieval, by GERE and three baselines, i.e., NSMN, DPR and RAG. 
Besides, we compare the memory footprint (disk space) needed by the overall GERE framework (including document and sentence retrieval) and the document retrieval baselines. 
The results are shown in Table~\ref{tab:memory}. 
GERE has a significant reduction of memory footprint and inference time of document retrieval. 
The major memory computation of GERE is a prefix tree of the document titles and the number of model parameters as opposed to a large document index and a dense vector for each document in existing works. 
These results suggest that our model is suitable for deployment in resource-limited platforms, e.g., the online verification device.

\section{Conclusion}
In this work, we proposed \textit{GERE} (for \textit{Generative Evidence REtrieval}), a novel generation-based framework to address the document retrieval and sentence retrieval jointly for FV. 
The generative formulation leads to several advantages with respect to current solutions, including improved time-efficiency and memory-efficiency, the ability to model document/sentence dependency, and a dynamic number of evidences.   
GERE can be easily adapted to existing claim verification models for better claim assessment. 
The experimental results on the FEVER dataset demonstrates the effectiveness of GERE.
In the future work, we would like to achieve the document retrieval and the sentence retrieval in a single decoder.  Besides, it is valuable to design an end-to-end FV system in a fully generative way.

\begin{acks}
This work was funded by the National Natural Science Foundation of China (NSFC) under Grants No. 62006218, 61902381, and 61872338, the Youth Innovation Promotion Association CAS under Grants No. 20144310, and 2021100, the Lenovo-CAS Joint Lab Youth Scientist Project, and the Foundation and Frontier Research Key Program of Chongqing Science and Technology Commission (No. cstc2017jcyjBX0059).
\end{acks}

\bibliographystyle{ACM-Reference-Format}
\bibliography{main}


\end{document}